\documentclass[journal]{IEEEtran}
\usepackage{amsmath,amsfonts}
\usepackage{algorithmic}
\usepackage{algorithm}
\usepackage{array}
\usepackage[caption=false,font=normalsize,labelfont=sf,textfont=sf]{subfig}
\usepackage{textcomp}
\usepackage{stfloats}
\usepackage{url}
\usepackage{verbatim}
\usepackage{graphicx}
\usepackage{cite}
\usepackage{color,array}
\usepackage{graphicx}
\usepackage{amsmath,amssymb,amsthm}
\usepackage{booktabs}
\usepackage{tikz}
\newtheorem{theorem}{Theorem}

\newtheorem{definition}{Definition}

\usepackage{tikz}
\usetikzlibrary{arrows.meta}
\usepackage[colorlinks,urlcolor=blue,linkcolor=blue,citecolor=blue]{hyperref}

\begin{document}

\title{Manifold Projection and Iterative Autoencoder Refinement for Masked Language Modeling}

\author{
Narges Mokhtari,
Farzan Haddadi\textsuperscript{*},
Ebrahim Rezaii%
\thanks{
\textsuperscript{*}Corresponding author (Email: farzanhaddadi@iust.ac.ir).
\\
School of Electrical Engineering, Iran University of Science \& Technology,
Tehran, Iran.\\
This work has been submitted to the IEEE for possible publication. Copyright may be transferred without notice, after which this version may no longer be accessible.
}
}

\maketitle

\begin{abstract}
In Transformer-based masked language models, attention is the primary mechanism for context mixing, but there are other ways to mix data across tokens. Recent attention-free mixers replace attention with fixed or hypernetwork-generated MLPs, alternating their dynamic, content-dependent weighting for computational simplicity.  
We build an alternative that gets the same property from a low-rank bottleneck autoencoder. 
We replace attention with a stack of autoencoder-based mixing modules, one operating over local neighborhoods, one over the full sequence, and one across attention heads, each compressing and reconstructing its input through a bottleneck, and its width is a hyperparameter rather than a training effect. 
In masked positions, we introduce an iterative refinement procedure that has two distinct steps. A pulling step that pulls an embedding representation toward a weighted average of its neighbors, and a correcting step that projects the result back to the learned manifold via an autoencoder.
Our architecture achieves a significant portion of attention's performance at about $1.9 \times$ fewer FLOPs when pretrained on C4 and evaluated with parameter-matched BERT baselines. 
Our model equals parameter-matched BERT and TinyBERT baselines on the rarest-token frequency bucket using a frequency-aware training schedule that samples rare tokens more than uniformly for the masking tasks.
\end{abstract}


\section{Introduction}

\IEEEPARstart{S}{ince} Vaswani in 2017\cite{vaswani2017attention} introduced self-attention as the Transformer's central context-mixing mechanism, the idea that combining data across tokens necessitates a dynamic, content-dependent weighting calculated from pairwise query-key comparisons has become standard in language modeling. The majority of advances in bidirectional text encoding are derived from masked language models based on this mechanism, such as BERT \cite{devlin2019bert}. Almost all of these models maintain self-attention even when they modify the pretraining objective, the parameter budget, or the training recipe around it.

There is a line of research that asks whether an attention mechanism is necessary at all. The MLP-Mixer \cite{tolstikhin2021mlpmixer}, gMLP \cite{liu2021pay}, FNet \cite{leethorp2022fnet}, and Synthesizer \cite{tay2021synthesizer} each replace self-attention with a fixed input-independent mixing operator. HyperMixer \cite{mai2023hypermixer} partially reintroduces input-dependence by generating mixing weights through a hypernetwork. These architectures demonstrate that attention is not strictly required for competitive performance.

Recent theoretical work shows that MLP-based mixers are already implicitly biased toward low-rank behavior, either through weight decay \cite{kuzborskij2025lowrank}, the dynamics of deep linear compositions \cite{jing2020irmae}, or through depth \cite{huh2022lowrank}. If this compression is happening regardless of architecture, leaving it implicit only makes it uncontrollable.
Now we ask a question: if content-dependent mixing does not require attention's query-key comparison, can it instead be built out of an explicit, controllable low-rank operation, whose compression ratio is a visible hyperparameter? This motivates us to use an autoencoder.

On the other hand, the dot product itself is known as Nadaraya--Watson kernel regression, a classical, similarity-weighted average \cite{nadaraya1964, watson1964, tsai2019transformer}. 
Separately, assuming a query as a noisy version of the true key and using Bayes' rule for the reconstruction shows that similarity-weighted averaging is the optimal estimator under that noise model, and simplifying that optimal estimator reproduces the softmax attention formula exactly \cite{bishop1995neural, henderson2022nvib}. We present our own second derivation based on our work notation in Section~\ref{Iterative refinement}.

\cite{haddadi2026diffusion} show that in attention, when only the $k$ nearest keys are effective under a small noise scale, the squared query--key distance splits, via the Pythagorean theorem, into a component along the local affine hull spanned by the neighboring keys and a component orthogonal to it; the orthogonal component cancels between the numerator and denominator of the softmax, so the attention output depends only on the query's orthogonal projection onto that affine hull, after which it returns a convex combination of the nearby keys, exactly the projection-then-reconstruction shape of an autoencoder.
They show the same projection behavior on the autoencoder side directly, with no noise assumption: a bottlenecked encoder's local null space, the directions along which the input can move without changing the code, and hence the output, has dimension $m-k$ for an $m$-dimensional input and $k$-dimensional code, so its orthogonal complement is exactly a $k$-dimensional locally estimated manifold, and any nearby off-manifold point is decoded by moving along this null space until it meets that manifold orthogonally.
As a result, attention and autoencoders can, in theory, be substituted. This is the approach we pursue here, creating an explicit autoencoder to replace attention.

We replace attention with a stack of autoencoders that perform the same similarity-weighted mixing role, but through a tunable bottleneck. We implement this principle at three levels: local neighborhoods (WindowMixAE), the full sequence (GlobalMixAE), and across attention heads (ChannelMixAE). We then extend it to masked-position prediction through an iterative refinement procedure.
Each refinement step has two steps: a similarity-weighted pulling step, which pulls an embedding toward a weighted average of its neighbors, and a correcting step, which pulls the result back toward the manifold of probable embeddings using the same autoencoder. 
Because this autoencoder is trained only on real, unmasked embeddings, we expect an off-manifold estimate projected back toward that same manifold (Section~\ref{Iterative refinement}).

\section{preliminaries}
\label{sec:preliminaries}

This section reviews the background on which our architecture is built on. Section~\ref{subsec:mlm} summarizes the masked language modeling paradigm, and Section~\ref{subsec:mixing} reviews the existing attention-free mixing mechanisms, which establish that self-attention can be replaced with an MLP network.
In Section~\ref{subsec:implicit-lowrank}, we demonstrate, based on recent theoretical results, that ordinary MLP-based mixers are implicitly biased toward low effective rank, regardless of whether their architecture makes the bottleneck explicit. 

\subsection{Masked Language Models}
\label{subsec:mlm}

Masked language modeling (MLM) has been the dominant pretraining paradigm for bidirectional text encoders since BERT \cite{devlin2019bert}. 
BERT masks 15\% of input tokens and predicts them from bidirectional context. Given a sequence $\mathbf{x} = [x_1, \dots, x_n]$, a random subset of positions  is replaced with a \texttt{[MASK]} token, and the model is trained to minimize
\begin{equation}
    \mathcal{L}_{\text{MLM}}(\theta) = \mathbb{E}_{\mathbf{x}} \sum_{i \in \mathcal{M}} -\log p_\theta(x_i \mid \mathbf{x}_{\text{masked}}),
\end{equation}
where $p_\theta(\cdot \mid \mathbf{x}_{\text{masked}})$ is the model's output distribution at position $i$ conditioned on the masked sequence. BERT is also trained with a Next Sentence Prediction (NSP) objective for sentence-pair understanding.

After BERT, RoBERTa \cite{liu2019roberta} was trained without any architectural changes on 10$\times$ more data with larger batches, removing NSP, dynamically remasking each epoch.
ALBERT \cite{lan2020albert} instead targets the parameter count, factorizing the $V\times H$ embedding matrix into $V\times E$ and $E\times H$ ($E\ll H$) and sharing Transformer block weights across layers, while replacing NSP with the harder Sentence-Order Prediction. 
ELECTRA \cite{clark2020electra} targets the efficiency of samples. ELECTRA trains a small generator to propose token replacements and a discriminator to classify every position as original or replaced.

BART \cite{lewis2020bart} extends the denoising method to a full encoder-decoder structure to generate rather than predict at masked positions. A bidirectional encoder reads text corrupted in different ways, such as individual tokens deleted, whole spans collapsed into a single mask, or sentences shuffled out of order. Then the autoregressive decoder is trained to reconstruct the original text.

Despite differences in the training objective, every model mentioned above has one structural property. The token-level hidden representation is passed through the network at constant width and constant sequence length, from the first layer to the last. 
ALBERT's factorized embedding is a low-rank decomposition of the vocabulary lookup table.  Once a token's embedding is projected to the model's hidden size $H$, every subsequent layer operates at that same full width, exactly as in BERT.

The first to explicitly decrease and then restore dimensions were Funnel-Transformer \cite{dai2020funnel} and Hourglass \cite{nawrot2022hierarchical}. 
Funnel-Transformer pools the hidden sequence to a shorter one as depth increases, such as spatial downsampling in a convolutional network, and then recovers the full length with a decoder stage for token-level objectives like MLM. Hourglass generalizes this into a symmetric, U-Net-style architecture. A contracting path downsamples the sequence, a matching expanding path restores it, and skip connections preserve fine-grained information that pooling would otherwise lose.

The key difference is which axis gets compressed. Funnel-Transformer and Hourglass compress along the sequence dimension, using pooling and Transformer sub-layers. Our mixing modules instead compress along the channel dimension (and, for GlobalMixAE, jointly across the sequence) using an autoencoder with a controllable bottleneck ratio, replacing self-attention entirely.

\subsection{Attention-Free Mixing Mechanisms}
\label{subsec:mixing}
A line of research replaces the self-attention layer with an ordinary, gradient-trained mixing operator, removing the dynamic, content-dependent weighting that attention provides.
MLP-Mixer~\cite{tolstikhin2021mlpmixer} and gMLP~\cite{liu2021pay}, both introduced in 2021, replace self-attention with a token-mixing MLP applied along the sequence dimension, whose weights are fixed after training rather than computed dynamically from the input. 
gMLP reports that its spatial gating unit matches Transformer perplexity on masked language modeling but underperforms on GLUE tasks that require cross-sentence relational reasoning (like MNLI), a gap the authors close by adding a small, single-head self-attention module (aMLP). 
FNet~\cite{leethorp2022fnet} replaces the token-mixing operation with an unparameterized 2D Fourier transform, eliminating learned mixing weights entirely. 
The Synthesizer~\cite{tay2021synthesizer} explores both dense and factorized input-independent attention weight matrices as drop-in replacements for self-attention, likewise fixing the mixing pattern after training.

The above methods fix their mixing weights once training is complete, but HyperMixer~\cite{mai2023hypermixer} generates the token-mixing MLP's weights dynamically via a hypernetwork applied to the input. The authors describe it as similar to the attention mechanism, because it reintroduces input-dependence into the mixing weights.

All of the methods above replace attention with a different single-pass mixing operator, computed once per layer. A separate line of work keeps the prediction itself iterative. Masked diffusion language models such as MDLM \cite{sahoo2024mdlm} recover masked tokens through repeated denoising steps rather than a single forward pass. MDLM's diffusion process operates over the discrete token distribution with a learned noise schedule, generating one masking level at a time. Our refinement loop operates directly in continuous embedding space with a fixed, deterministic predictor-corrector update and no noise schedule.

Our GlobalMixAE module applies a transform along the token dimension with weights shared across channels. It imposes an explicit low-rank bottleneck rather than an unconstrained $n \times n$ mixing matrix. Also, it is one stage of a local-then-global-then-cross-head hierarchy (WindowMixAE, GlobalMixAE, ChannelMixAE) followed by an iterative refinement stage, rather than the sole mixing operation applied once per layer.

\subsection{Low-Rank Structure in MLPs}
\label{subsec:implicit-lowrank}

In an autoencoder, the encoder compresses its input to a fraction $r<1$ of its original width before the decoder expands it back. MLPs have no such bottleneck, and their weight matrices are numerically full-rank, but their effective rank in practice is less than algebraic rank. Three independent lines of evidence support this observation, and each has the same consequence for mixing-based architectures. 

\subsubsection{Measuring effective rank}

The algebraic rank is not informative for a trained matrix, since noise can cause the weight matrix to be full rank. Following \cite{kuzborskij2025lowrank} (building on \cite{timor2023implicit}), we instead use the stable rank.

\begin{definition}[Stable rank]
\label{def:srank}
For $W \in \mathbb{R}^{m\times n}$ with singular values $\sigma_1 \ge \cdots \ge \sigma_r \ge 0$, $r = \min(m,n)$,
\begin{equation}
\operatorname{srank}(W) \;=\; \frac{\|W\|_F^2}{\|W\|_2^2}
\;=\; \frac{\sum_i \sigma_i^2}{\sigma_1^2}
\;\in\; [1, r].
\label{eq:srank}
\end{equation}
Here $\|W\|_F$ is the Frobenius norm, the square root of the sum of all squared entries of $W$, and $\|W\|_2$ is the spectral norm, simply the largest singular value $\sigma_1$.
\end{definition}
A uniform singular spectrum gives $\operatorname{srank}(W)=r$. A spectrum dominated by one direction drives it towards $1$, regardless of how many singular values are technically nonzero.

\cite{kuzborskij2025lowrank} show that $L_2$ weight decay provably forces the stable rank downward. At any stationary point of the regularized loss, every layer's Frobenius norm is equal regardless of width, so a wide layer does not get an extra energy budget to spread across its extra directions.

\begin{theorem}[Low-rank bias under weight decay, \cite{kuzborskij2025lowrank}]
\label{thm:kuzborskij}
Let $\theta$ be a stationary point of the $L_2$-regularized loss $L_\lambda$ for a $K$-layer network. Then
\begin{equation}
\frac{1}{K}\sum_{k=1}^K \operatorname{srank}(W_k)^{-1/2}
\;\ge\; \left(\frac{\lambda}{L(\theta)^{1/2-1/K}}\right)^{1/2}.
\label{eq:kuzborskij-bound}
\end{equation}
The stronger the weight decay (the larger $\lambda$), the lower the layers' stable ranks. For deep linear networks, this pressure is strong enough that every single weight matrix eventually converges to exactly rank one.
\end{theorem}

\subsubsection{Rank-Minimizing Autoencoder}
\cite{jing2020irmae} designed an experiment. It inserted a stack of $K$ extra fully linear layers $W = W_K W_{K-1}\cdots W_1$ between the encoder and the decoder, without a bottleneck, without a rank penalty, and without a change in reconstruction loss. Trained by gradient flow under a balanced initialization, the end-to-end matrix $W(t)$ is known to evolve as
\begin{equation}
\resizebox{\linewidth}{!}{$\displaystyle
\dot W(t) = -\sum_{k=1}^K \big(W(t)W(t)^\top\big)^{\frac{k-1}{K}}\,
\nabla L^1(W(t))\,\big(W(t)^\top W(t)\big)^{\frac{K-k}{K}}
$}
\label{eq:balanced-flow}
\end{equation}
where $L^1$ is the reconstruction loss as a function of the end-to-end matrix. Writing $W(t)=U(t)\Sigma(t)V(t)^\top$ for its SVD, Equation~(\ref{eq:balanced-flow}) determines the dynamics of each singular value.

\begin{theorem}[Rich-get-richer singular value dynamics; \cite{jing2020irmae}, building on \cite{arora2018implicit}]
\label{thm:irmae}
Under gradient flow with balanced initialization, for a stack of $K$ extra linear layers, each singular value $\sigma_i(t)$ of the end-to-end map evolves with relative growth rate $\dot\sigma_i/\sigma_i \propto \sigma_i^{1-2/K}$. Since $1-2/K>0$ for $K>2$, larger singular values grow relatively faster than smaller ones, concentrating the spectrum on a few directions.
\end{theorem}
Once one direction is even slightly larger, it grows faster than the rest during training (a fact that \cite{jing2020irmae} confirm experimentally).

\subsubsection{Regardless of training mechanism}
The third research is independent of the training process \cite{huh2022lowrank}. The bias toward low effective rank exists in the parameterization itself, before any training. Since $\operatorname{rank}(AB) \le \min(\operatorname{rank}(A), \operatorname{rank}(B))$, depth by itself can only lower rank and never raise it. They used a second metric to formalize this. 

\begin{definition}[Effective rank; \cite{roy2007effective}]
\label{def:effrank}
For $A \in \mathbb{R}^{m\times n}$ with normalized singular values $\bar\sigma_i = \sigma_i/\sum_j\sigma_j$,
\begin{equation}
\rho(A) \;=\; -\sum_{i=1}^r \bar\sigma_i \log(\bar\sigma_i).
\label{eq:effrank}
\end{equation}
\end{definition}
This is similar to stable rank in practice, both are maximized when singular values are all equal.

\begin{theorem}[Rank contraction with depth, \cite{huh2022lowrank}]
\label{thm:huh}
For a $d$-layer linear network with i.i.d.\ Gaussian weight matrices, in the limit $\dim(W)\to\infty$,
\begin{equation}
\rho\big(W_d W_{d-1} \cdots W_1\big) \;\le\; \rho\big(W_{d-1}\cdots W_1\big).
\label{eq:huh-bound}
\end{equation}
Stacking one more Gaussian layer never increases effective rank, and generically decreases it.
\end{theorem}

As a result, MLPs are biased toward low effective rank whether or not their architecture makes a bottleneck explicit. Every attention-free mixer introduced in Section~\ref{subsec:mixing}, MLP-Mixer's token-mixing MLP, gMLP's spatial gating unit, and FNet's linear token mixer, is already being pushed toward low effective rank.
Given this, if MLP-based mixing is already compressing to low rank regardless of architecture, leaving that compression implicit only makes it uncontrolled. We instead build the mixing operation directly out of an autoencoder with an explicit bottleneck Section~\ref{sec:method}), whose compression ratio $r$ is a visible, tunable hyperparameter.

\section{Method}
\label{sec:method}

\begin{figure}[htbp]
\centering
\includegraphics[width=0.8\linewidth]{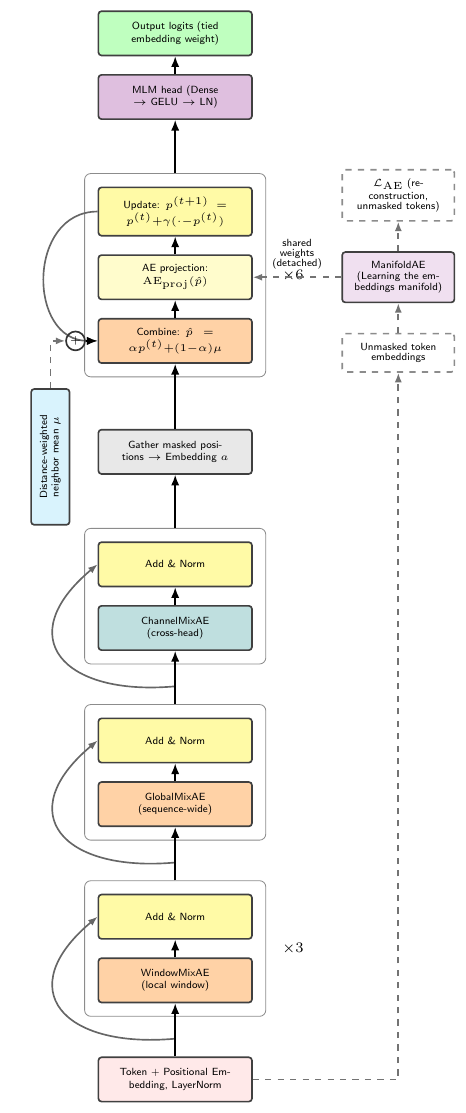}
\caption{The model architecture}
\label{fig:architecture}
\end{figure}

\begin{figure}[htbp]
\centering
\includegraphics[width=1.1\linewidth]{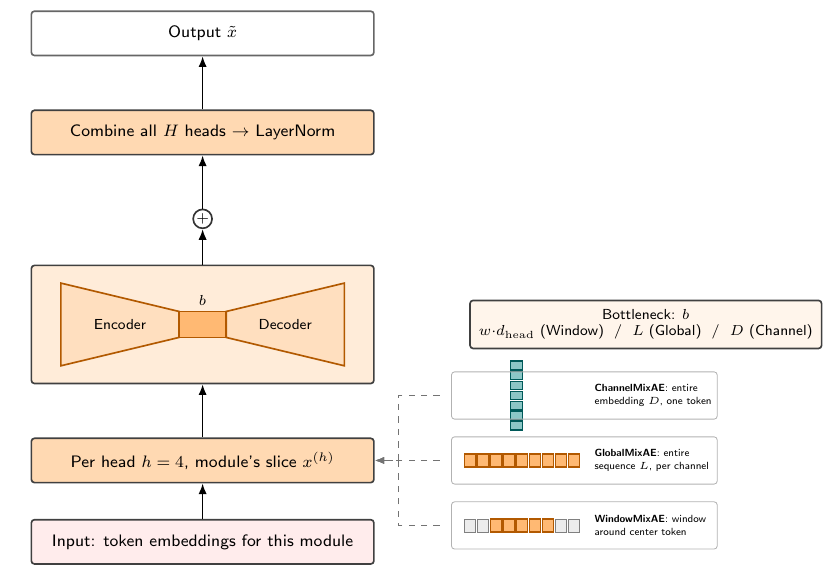}
\caption{Internal structure of the mixing modules.}
\label{fig:internal}
\end{figure}

Our model follows the standard three-stage encoder pipeline of a Transformer-based masked language model, an embedding stage, a context mixing stage, and a task-specific output stage (a masked-token prediction head during pretraining and a pooled classification head during GLUE fine-tuning), but replaces the self-attention layers with a hierarchy of autoencoder-based mixing modules, followed by an iterative refinement stage applied only at masked positions. Fig.~\ref{fig:architecture} and Fig.~\ref{fig:internal} demonstrate the model architecture and internal structure of the mixing module, respectively. The token embeddings are processed by a stack of local mixing layers (WindowMixAE), a global mixing layer (GlobalMixAE), and then a ChannelMixAE layer, which mixes information across independent heads. The resulting per-position representations at masked positions are taken as an initial embedding and iteratively refined by a distance-weighted average of unmasked token embeddings and projected through a manifold autoencoder (ManifoldAE).

\subsection{Local mixing stage: WindowMixAE}
The model uses a multihead AE structure, which means that the Autoencoders process $d = D/H$ ($H=4$ heads) of each embedding dimension. 
For a token at position $i$ with embedding $x_i \in \mathbb{R}^{D}$, a window of $w=5$ neighboring embeddings, positions $i-2,\dots,i+2$, is concatenated per head and passed through an encoder to a bottleneck of dimension $b = d/2$, then decoded back to $d$ dimensions and added as a residual to the center token's embedding:
\begin{equation}
  \tilde{x}_i^{(h)} = x_i^{(h)} + \mathrm{Dec}^{(h)}\!\left(\mathrm{Enc}^{(h)}\!\left(\left[x_{i-2}^{(h)}; \dots; x_{i+2}^{(h)}\right]\right)\right),
\end{equation}
where superscript $(h)$ denotes the per-head slice. Three such layers are stacked, giving an effective local receptive field of $\pm 6$ tokens.

\subsection{Global mixing stage: GlobalMixAE}
Following the local mixing stack, a single GlobalMixAE layer mixes information across the entire sequence. For channel $c \in \{1,\dots,d\}$ within head $h$, let
\begin{equation}
  v_c^{(h)} = \big(x_{1,c}^{(h)}, \dots, x_{L,c}^{(h)}\big) \in \mathbb{R}^{L}
\end{equation}
denote the sequence of values that the channel takes across all $L$
positions. This vector is passed through an encoder/decoder pair,
with weights shared across all $d$ channels of a head, to a
bottleneck of dimension $b_G = L/4$, and added back as a residual:
\begin{equation}
  \tilde v_c^{(h)} = v_c^{(h)} + \mathrm{Dec}_G^{(h)}\!\left(\mathrm{Enc}_G^{(h)}\!\left(v_c^{(h)}\right)\right),
\end{equation}

\subsection{Channel mixing stage: ChannelMixAE}
Since each head operated on an independent $d = D/H$ dimensional slice of the model dimension, only the final dense layer sees the full concatenated $D$-dimensional vector. 
We introduce an additional ChannelMixAE module, applied after GlobalMixAE, that
uses the full embedding $x_i \in \mathbb{R}^D$ of each token as a single vector, not split across heads, and passes it through an encoder/decoder pair to a bottleneck of dimension $b_C =D/2$, added back as a residual delta:
\begin{equation}
  \tilde{x}_i = x_i + \mathrm{Dec}_C\!\left(\mathrm{Enc}_C(x_i)\right),
\end{equation}
where $\mathrm{Enc}_C$ and $\mathrm{Dec}_C$ are MLPs whose
intermediate widths follow the same geometric taper used by
WindowMixAE. This is the only point in the mixing
pipeline where information from different heads is combined before
the final classification/output layer. Like all other modules in
this architecture, ChannelMixAE performs no dot product and remains fully attention-free.

\subsection{Iterative refinement}
\label{Iterative refinement}

The mixing modules combine information across tokens and channels. Our model goes further by adding a refinement stage that operates on the embeddings produced by the mixing modules.
The refinement stage relies on a network trained on the unmasked tokens separately, which is ManifoldAE. At every training step, we take the embeddings of all unmasked tokens in the batch and train this autoencoder, with its own reconstruction loss. 
Once trained, the iterative refinement loop reuses this same encoder-decoder, with weights frozen (no gradients flow into it from this second use) and has two steps:
\begin{enumerate}
    \item Pulling Step: The masked position embedding produced by the mixing module moves one step toward a distance-weighted average of neighboring (unmasked) embeddings. Because this average is a combination of points on a curved manifold, it generally lands off that manifold. So by moving one step toward it, the embedding gets off the manifold.
    \item Correcting step: Because the autoencoder was trained on unmasked embeddings and knows the embedding manifolds, we expect that by passing the off-manifold vector through an autoencoder, it will be projected back to the manifold of real embeddings.
    As discussed earlier, \cite{haddadi2026diffusion} give this expectation a structural, noise-free grounding: a bottlenecked autoencoder's local null space around each training point is exactly what defines this pull toward the manifold it was trained to represent.
\end{enumerate}

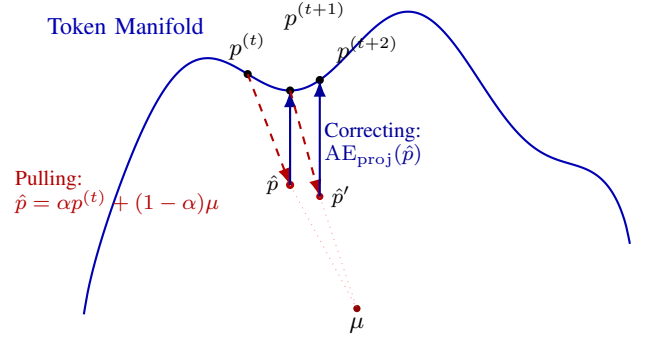
\begin{figure}[t]
\centering
\begin{tikzpicture}[scale=0.85, every node/.style={font=\small}]

  \draw[thick, blue!70!black, domain=-4.15:4.35, samples=150, smooth]
    plot ({\x}, {sqrt(4.3*4.3 - (\x-0.1)*(\x-0.1)) + 0.6*sin(100*(\x-0.1))});

  \node[blue!70!black] at (-3.5, 4.6) {Token Manifold};

  \coordinate (P0) at (-1.6, 3.845);
  \fill[black] (P0) circle (1.8pt);
  \node[above=2pt] at (P0) {$p^{(t)}$};

  \coordinate (MU) at (0.1, 0.2);
  \fill[red!55!black] (MU) circle (1.6pt);
  \node[below] at (MU) {$\mu$};

  \coordinate (PH) at (-0.94, 2.122);
  \fill[red!70!black] (PH) circle (1.8pt);
  \node[left=1pt, font=\footnotesize] at (PH) {$\hat p$};

  \draw[-{Latex[length=2.2mm]}, red!70!black, thick, dashed]
    (P0) -- (PH);
  \node[red!70!black, font=\footnotesize, align=left, anchor=east]
    at (-1.9, 2) {Pulling:\\ $\hat p=\alpha p^{(t)}+(1-\alpha)\mu$};

  \draw[red!40, thin, dotted] (PH) -- (MU);

  \coordinate (P1) at (-0.94, 3.59);
  \fill[black] (P1) circle (1.8pt);
  \node at (-0.55, 4.75) {$p^{(t+1)}$};
  \draw[-{Latex[length=2.2mm]}, blue!60!black, thick]
    (PH) -- (P1);
  \node[blue!60!black, font=\footnotesize, align=left, anchor=west]
    at (-0.55, 2.75) {Correcting:\\ $\mathrm{AE}_{\mathrm{proj}}(\hat p)$};

  \coordinate (PH2) at (-0.478, 1.943);
  \fill[red!70!black] (PH2) circle (1.6pt);
  \node[right=1pt, font=\footnotesize] at (PH2) {$\hat p'$};

  \draw[-{Latex[length=2.2mm]}, red!70!black, thick, dashed]
    (P1) -- (PH2);
  \draw[red!40, thin, dotted] (PH2) -- (MU);

  \coordinate (P2) at (-0.478, 3.753);
  \fill[black] (P2) circle (1.8pt);
  \node[above right=3pt] at (P2) {$p^{(t+2)}$};
  \draw[-{Latex[length=2.2mm]}, blue!60!black, thick]
    (PH2) -- (P2);

\end{tikzpicture}
\caption{The iterative refinement procedure consists of two steps: pulling and correcting. Starting from $p^{(t)}$, the pulling step moves the embedding toward the neighbor mean $\mu$, landing off the manifold at $\hat p$. Then in the correcting step $\mathrm{AE}_{\mathrm{proj}}(\hat p)$ project this back onto the manifold, giving $p^{(t+1)}$.}
\label{fig:predictor-corrector}
\end{figure}

Fig.~\ref{fig:predictor-corrector} demonstrates this process. 
By iteratively repeating this pulling step and correcting step, we refine the embedding of masked positions. We state this observation empirically in ablation. The refinement loop performs the following update formula per iteration:

\begin{align}
\hat p &= \alpha\, p^{(t)} + (1-\alpha)\, \mu,
&&\text{Pulling Step}
\label{eq:predictor} \\
p^{(t+1)} &= p^{(t)} + \gamma\big(\mathrm{AE}_{\mathrm{proj}}(\hat p) - p^{(t)}\big),
&&\text{Correcting Step}
\label{eq:corrector}
\end{align}
where $\gamma$ controls the step size, and $\mu \in \mathbb{R}^D$ is the distance-weighted mean over unmasked token embeddings,
\begin{equation}
  \mu = \sum_{j \in \mathcal{K}} \mathrm{softmax}_j\!\left(-\frac{|i-j|}{\tau}\right) x_j,
  \label{eq:neighbor-mean}
\end{equation}

with $\mathcal{K}$ the set of unmasked positions and $\tau$ the temperature of this softmax. For large $\tau$, the weights become nearly uniform, causing $\mu$ to approach the global mean. In contrast, small $\tau$ places most of the weight on the nearest token, while intermediate values of $\tau$ lead to stable states, where $\mu$ converges toward a small cluster of similar tokens. During training, $\tau$ is annealed from $\tau_{\min}=0.5$ to $\tau_{\max}=12$ according to an inverse-cosine schedule.

Let $a \in \mathbb{R}^{D}$ be the initial embedding of the masked position (the output of the local, global, and channel mixing stages). Starting from $p^{(0)} = a$, the refinement loop applies the pulling-correcting update of Equation~(\ref{eq:predictor}--\ref{eq:corrector}) for $T=6$ fixed iterations, with $\alpha=0.7$ and step size $\gamma=0.4$. Here $\mathrm{AE}_{\mathrm{proj}}$ is a forward pass through the ManifoldAE with its weights detached from the refinement loop's gradient. 

Only the ManifoldAE's own parameters are detached, so no gradient updates flow into them from the refinement loop. The forward computation itself is not detached from its input. $\mathrm{AE}_{\mathrm{proj}}$ is used as an ordinary differentiable function of $\hat p$, with its weights treated as constants inside that computation. Gradients therefore do flow back through $\mathrm{AE}_{\mathrm{proj}}$'s layers to $\hat p$, and from there to earlier refinement iterations and the mixing modules; this is exactly what makes the vanishing-gradient failure mode described above possible, and what the damped residual update was designed to fix.

The residual form of this update, with $\gamma<1$, turned out to be essential. We first tried the more direct update $p^{(t+1)} \leftarrow \mathrm{AE}_{\mathrm{proj}}(\hat p)$, replacing $p^{(t)}$ outright at every step. Because this replacement is applied six times in a row, and each pass through the ManifoldAE is itself several layers deep, the effective path from the last refinement step back to the first is comparable to an unusually deep network with no skip connections. In practice, this caused the gradient to vanish: by the time it reached the first iteration, its norm was effectively zero. Switching to the damped residual update above fixed this. With everything else held fixed, the ratio between the gradient norm at the first iteration and at the last rose from $\approx 0$ to $\approx 0.446$. The earliest refinement steps started receiving a meaningful learning signal again.

\subsection{Retrieval as Bayesian Denoising}
\label{sec:kernel-derivation}

Considering attention as the result of Bayesian denoising under Gaussian corruption is not a new observation. \cite{bishop1995neural} showed, in the context of ordinary classification rather than attention, that if a set of classes has Gaussian feature distributions with equal covariance, the posterior class probabilities computed via Bayes' rule take the exact form of a softmax over linear discriminants, reached by asking a completely different question (which class does this point belong to, not which key does this query match).
\cite{henderson2022nvib} treat a query as a Gaussian-corrupted copy of the key to be retrieved, and show that applying Bayes' rule to recover the most likely key reconstructs softmax attention exactly.

We follow this derivation ourselves in our own notation, because the same reasoning that a noisy input needs to be projected back toward the known manifold motivates what the ManifoldAE is doing in Section~\ref{Iterative refinement}.

Attention compares a query against a set of keys and returns a
weighted blend of the associated values. Suppose each key $k_i$ is
clean, and a query $q_j$ is a noisy probe of one of them,
\begin{equation}
q_j = k_i + \eta, \qquad \eta \sim \mathcal{N}(0, \sigma^2 I).
\label{eq:noisy-query}
\end{equation}
We want $P(k_i \mid q_j)$, but can only compute the reverse,
$P(q_j \mid k_i)$, directly from this noise model. Bayes' rule bridges
the two:
\begin{equation}
\underbrace{P(k_i \mid q_j)}_{\text{want this}} =
\frac{\overbrace{P(q_j \mid k_i)}^{\text{can compute this}}\, P(k_i)}{P(q_j)}.
\label{eq:bayes}
\end{equation}
With Gaussian $P(q_j\mid k_i) \propto \exp(-\lVert q_j-k_i\rVert^2/2\sigma^2)$
and a uniform prior $P(k_i)=1/N$, every other factor is constant in
$i$ (fixed once $q_j$ is observed), so
\begin{equation}
P(k_i \mid q_j) \;\propto\;
\exp\!\left(-\frac{\lVert q_j - k_i \rVert^2}{2\sigma^2}\right).
\label{eq:soft-assignment}
\end{equation}
The best single value $z$ to report is the one minimizing expected
squared error, $\sum_i P(k_i\mid q_j)\lVert z-v_i\rVert^2$; setting its
gradient to zero gives the weighted mean
\begin{equation}
z^\ast = \frac{\sum_i P(k_i \mid q_j)\, v_i}{\sum_i P(k_i \mid q_j)},
\label{eq:weighted-mean}
\end{equation}
the Nadaraya--Watson estimator. Expanding
$\lVert q_j-k_i\rVert^2 = \lVert q_j\rVert^2 - 2\langle q_j,k_i\rangle + \lVert k_i\rVert^2$,
the first term is constant in $i$ and cancels; the last term also
cancels once keys have approximately equal norm, $\lVert
k_i\rVert\approx c$ (after normalization). What remains is
exactly the softmax attention,
\begin{equation}
z^\ast = \mathrm{softmax}\!\Big(\tfrac{q_j K^\top}{\sigma^2}\Big) V,
\label{eq:final-attention}
\end{equation}
which matches the standard $\mathrm{softmax}(q^\top k/\sqrt{d_k})$
scaling of \cite{vaswani2017attention} when $\sigma^2 =
\sqrt{d_k}$. A large $\sigma$ makes the weights in Equation~(\ref{eq:soft-assignment} nearly
uniform, averaging over most keys, while a small $\sigma$ concentrates
almost all the weight on the single nearest key, the same bandwidth
trade-off that governs the temperature $\tau$ in our own neighbor
mean $\mu$.
\cite{haddadi2026diffusion} study this convergence pattern
for the same closed-form estimator: they show, and confirm with their
own simulation, that as $\sigma$ is gradually decreased the estimator
moves from an average over nearby signals toward a sparse solution
concentrated on the single nearest one. The same qualitative behavior
is expected to hold for our own $\tau$.

\begin{table*}[!t]
\centering
\caption{Comparison across AE-mixing configurations, all on C4, 100k steps. Bucket columns report masked-token accuracy (\%) for frequency ranges $[100,2\mathrm{k})$, $[2\mathrm{k},5\mathrm{k})$, and
$[5\mathrm{k})$.}
\label{tab:ae-variants}
\resizebox{\linewidth}{!}{%
\begin{tabular}{@{}lrrrrrrrrr@{}}
\toprule
Config.\ & $D$ & $L$ & Params (M) & Time/step & Score\ (\%) & [100,2k) & [2k,5k) & [5k) & PPL \\
\midrule
Base                              & 680 & 512 & 32   & 51s  & 	38.1          & 51.2 & 37.3         & 23.4          & 33.1 \\
Base ($D{=}384$)                  & 384 & 512 & 16.8 & 39s   & 34.4          & 47.4 & 33.4          & 19.9          & 41.4 \\
B: ChunkedContentBias              & 680 & 512 & 32   & 59s & \textbf{41.2} & \textbf{58.3} & \textbf{38.9}          & \textbf{24.9} & \textbf{27.4} \\
C: Extended training dynamics & 680 & 128 & 31   & 52s  & 	37.8          & 50.6 & 37.4          & 22.2          & 36.6 \\
\bottomrule
\end{tabular}%
}
\end{table*}

\subsection{Experimental Setup \& Configuration}
The model is trained with a masked language modeling objective using 15\% masked tokens. It is applied in two phases that differ in how the masked-token target distribution and the loss are weighted with respect to token frequency.
Phase 1 (steps 1--40k) was standard, frequency-uniform MLM, matching the original BERT masking procedure \cite{devlin2019bert}. For the first 40k steps, the masked targets are uniformly at random from the tokens in each sequence, and the cross-entropy loss on every masked token is weighted equally. 
Phase 2 (steps 40k--100k) was frequency-aware masking and loss weighting. Instead of uniform sampling, masking targets are drawn with a probability that increases with token rarity, so that rare tokens are selected as masking targets more often than common tokens. Moreover, independently of how a masked token was selected, its contribution to the cross-entropy loss is scaled by a factor that depends on its frequency bucket. This mechanism controls how much a prediction error on the selected token matters to the gradient.
Independent of this two-phase schedule, validation always uses the uniform masking distribution with unweighted loss, so that the reported validation metric remains comparable with different architectural configurations evaluated throughout this work.

Training configuration and number of tokens used during training are important factors in determining whether the attention-free architecture becomes competitive with self-attention mechanisms.
We evaluate the performance of our model using different system configurations. 
For the base configuration (A) described in Section~\ref{sec:method}, we evaluate the performance of our model using the 10B-token C4 corpus \cite{raffel2020exploring}. The base configuration is four stacked WindowMixAE layers, a single GlobalMixAE layer, and a ChannelMixAE layer, and then the iterative refinement loop with six iterations.
The model has a hidden dimension $D=680$, sequence length $L=512$, and 4 mixing heads, for a total of 32M parameters. Training uses a batch size of 128 for 100{,}000 steps with the AdamW optimizer, a cosine learning-rate schedule and mixed-precision (fp16) training, using the NVIDIA L40S GPU with 48 GB of GDDR6 memory.

Motivated by HyperMixer \cite{mai2023hypermixer}, Configuration B designed a Content-aware neighbor bias. 
The neighbor-averaging term $\mu$ in the refinement loop weights known tokens solely by their distance to the masked position, independent of token content.
ChunkedContentBias computes a content-dependent bias between each masked position's mixing-derived embedding and the chunked mean of the sequence, which is then added to the distance-based logits before the softmax that produces $\mu$.
The sequence is partitioned into fixed-size windows of $\texttt{CONTENT\_CHUNK\_SIZE}=8$. Each chunk is summarized by the mean embedding of its known tokens, and scored against the masked position's own embedding via cosine similarity, giving a content-dependent bias added to the distance-based logits for every token in that chunk.

In Configuration C, the two mixing/refinement coefficients $\alpha$ and $\gamma$, previously fixed, were instead made learnable. In
addition, an exponential moving average (EMA) of the model weights
was tracked and used during evaluation. The training procedure was
also modified to incorporate gradient accumulation.

The ManifoldAE is a multi-head autoencoder: the $D$-dimensional embedding is split into $H=4$ independent heads of width $d_{\mathrm{head}} = D/H$, and each head has its own 3-hidden-layer encoder-decoder stack with a bottleneck ratio of $r=0.5$ (bottleneck width $= 0.5 \cdot d_{\mathrm{head}}$), GELU-activated, with intermediate widths following a geometric taper -- the same hourglass template used by WindowMixAE and GlobalMixAE. All models are trained
with AdamW ($\beta_1{=}0.9$, $\beta_2{=}0.98$, $\epsilon{=}10^{-6}$, weight decay $0.01$ applied to all two-or-more-dimensional parameters except embeddings, layer norms, and biases), a cosine learning-rate schedule with linear warmup, and mixed-precision (fp16) training.

The ManifoldAE's reconstruction loss is combined with the main MLM loss at a weight that is itself annealed over training, from $0.3$ at initialization down to $0.09$ (30\% of the initial value) by the end of training, following a cosine schedule. We found this necessary: holding the weight fixed at $0.3$ throughout let the ManifoldAE's reconstruction loss begin rising again partway through training, most likely because the MLM loss increasingly competes for optimization budget as training progresses; annealing the AE weight down gives the MLM objective more room in the second half of training without destabilizing the ManifoldAE early on, when it most needs to learn the embedding manifold.

\begin{table*}[!t]
\centering
\caption{Our AE base configuration versus three BERT-based baselines,
matched in parameter count. \textsuperscript{*}BERT here is standard self-attention Transformer encoders trained from scratch under our own setup (same optimizer, masking scheme, and steps as our model).}
\label{tab:vs-baselines-pretrain}
\resizebox{\linewidth}{!}{%
\begin{tabular}{@{}lrrrrrrrrrr@{}}
\toprule
Model & $D$ & $L$ & Batch & Params (M) & Time/step & Score\ (\%) & [100,2k) & [2k,5k) & [5k) & PPL \\
\midrule
\textbf{Our model} & 680 & 512 & 128 & 32    & 51s  & 38.1 & 51.2 & 37.3 & 23.4 & 33.1 \\
TinyBERT             & 384 & 128 & 192 & 31    & 61s  & 43.1 & 70.2  & 40.9 & 22.0 & 25.8\\
BERT\textsuperscript{*}                & 680 & 512 & 128 & 28 & 65s  & 46.6 & 71.9& 42.1 & 24.4 & 20.9 \\
BERT\textsuperscript{*}               & 680 & 128 & 192 & 32    & 71s  & 46.9 & 74.0 & 44.3 & 24.8 & 18.0 \\
\bottomrule
\end{tabular}%
}
\end{table*}

\begin{table*}[!t]
\centering
\small
\caption{GLUE validation results over 3 seeds: our best AE configuration (base $+$
ChunkedContentBias) versus the three attention-based baselines of
Table~\ref{tab:vs-baselines-pretrain}.}
\label{tab:vs-baselines-glue}
\resizebox{\linewidth}{!}{%
\begin{tabular}{@{}lrrrrr@{}}
\toprule
Task & Metric & AE (+ContentBias) & Attention ($D{=}384,L{=}128$) & Attention ($D{=}680,L{=}128$) & Attention ($D{=}680,L{=}512$) \\
\midrule
CoLA  & Matthews corr.       & \textbf{25.9$\pm$1.5} & 17.7 & 19.0 & 21.9$\pm$1.3 \\
SST-2 & Accuracy             & \textbf{85.5$\pm$0.8} & 80.6 & 84.0 & 84.2$\pm$0.6 \\
MRPC  & Accuracy / F1        & 69.5$\pm$0.7 / 79.9$\pm$0.3 & 72.4 / 80.9 & 73.7 / 81.8 & 82.1$\pm$0.8 / 87.1$\pm$1.2 \\
STS-B & Pearson / Spearman   & 65.5$\pm$0.9 / 65.3$\pm$1.1 & 79.9 / 79.6 & 82.4 / 82.1 & 83.0$\pm$0.4 / 82.8$\pm$0.4 \\
QQP   & Accuracy / F1        & \textbf{79.8$\pm$0.2} / \textbf{73.8$\pm$0.8} & \textbf{63.1} / \textbf{60.2} & 83.9 / 78.4 & 83.3$\pm$0.1 / 76.7$\pm$0.4 \\
MNLI  & Matched / Mismatched & \textbf{53.2$\pm$0.6} / \textbf{53.8$\pm$0.8} & \textbf{31.8} / \textbf{31.8} & 66.2 / 67.1 & 63.7$\pm$0.1 / 64.7$\pm$0.2 \\
QNLI  & Accuracy             & \textbf{74.0$\pm$0.1} & \textbf{50.5} & 80.6 & 80.8$\pm$0.4 \\
RTE   & Accuracy             & 52.7$\pm$4.1 & 53.3 & 55.6 & 57.0$\pm$2.7 \\
WNLI  & Accuracy             & 28.1$\pm$5.0 & 28.1 & 33.8 & 36.1$\pm$6.6 \\
\bottomrule
\end{tabular}%
}
\end{table*}

\section{Results}
\label{sec:results}
We report Score, top-1 masked-token prediction accuracy averaged over all masked positions in the validation set, regardless of token frequency. The bucket columns ([100,2k), [2k,5k), [5k)) report the same accuracy restricted to masked tokens whose frequency rank is in this range.

All four configurations, pretrained under the same circumstances on C4 for 100k steps, are compared in Table~\ref{tab:ae-variants}. It is confirmed that allowing the neighbor-averaging term to depend on content rather than simply distance is worth the additional parameters by adding ChunkedContentBias (Configuration B), which causes the best overall result in both accuracy and perplexity.

Our base AE configuration is compared to BERT and TinyBERT Transformer baselines with comparable parameter counts in Table~\ref{tab:vs-baselines-pretrain}. 
The attention baselines are standard Transformer encoders matched in hidden dimension $D$ and parameter count to our configuration: the $D{=}680, L{=}512$ baseline uses 2 layers and 4 attention heads, and the $D{=}680, L{=}128$ baseline uses 4 layers and 8 attention heads. 
Full configuration details for all baselines are available in the released code (\href{to be added}{GitHub}).

Table~\ref{tab:vs-baselines-glue} fine-tunes our model configuration (base $+$ ChunkedContentBias) and all BERT baselines on GLUE. The AE model outperforms three attention baselines on CoLA and SST-2. In tasks that require relating two input segments to each other, such as MRPC, the mixing modules perform poorly. CoLA and SST-2 are both single-sentence tasks, solvable from one segment's surface-level or lexical properties, whereas the tasks where the AE model falls behind require judging a relationship between two separate spans of text. The token-to-token alignment mechanism missing from our architecture is exactly what self-attention computes, a pairwise, content-dependent weighting between every pair of tokens, at a substantial computational cost. 
In contrast, our own approach, especially in its base configuration, just considers the physical distance between neighboring tokens rather than their semantic content. Nevertheless, by this simple positional average, could fill a masked token semantically correctly.

\begin{table}[!t]
\centering
\small
\caption{FLOPs per forward pass, base + ChunkedContentBias configuration (batch size 32).}
\label{tab:compute-cost-contentbias}
\begin{tabular}{@{}lrr@{}}
\toprule
Configuration & FLOPs & FLOPs \\
 & ($L{=}128$) & ($L{=}512$) \\
\midrule
\textbf{Full model} & \textbf{114.7 G} & \textbf{477 G} \\
\quad WindowMixAE $\times3$    & 63.95 G & 255.79 G \\
\quad GlobalMixAE               & 1.40 G  & 22.36 G  \\
\quad ChannelMixAE              & 16.3 G  & 65.3 G  \\
\quad ManifoldAE reconstruction (training pass) & 3.47 G  & 13.85 G  \\
\quad ChunkedContentBias        & 0.10 G  & 0.41 G   \\
\quad Refinement loop ($T{=}6$) & 3.63 G  & 14.71 G  \\
\quad MLM output head           & 25.80 G & 104.56 G \\
\midrule
Matched attention & 218.3 G & 907.6 G \\
\bottomrule
\end{tabular}
\end{table}

Beyond the accuracy comparison above, the two architectures are different structurally in how their cost scales with sequence length $L$.
Self-attention costs $O(L^2D)$ per layer, dominated by the $QK^\top$ and softmax-weighted-value calculation terms. Our mixing modules instead use fixed-size autoencoder bottlenecks, so most of them cost only $O(LD)$, linear, not quadratic, in sequence length $L$. 
The one exception is GlobalMixAE, whose bottleneck width scales with $L$ and so keeps a (smaller-constant) $O(L^2)$ term.
Table~\ref{tab:compute-cost-contentbias} presents FLOPs at the two lengths we trained and evaluated. At both $L=128$ and $L=512$, our strongest configuration, which uses a content-dependent mean, needs $1.9\times$ fewer FLOPs than a matched attention model, even though it utilizes an architecture that is by design less capable of modeling paired token interaction than attention.

\begin{table}[!t]
\centering
\footnotesize
\caption{Ablations over the refinement loop's three hyperparameters: number of iterations $T$, mixing ratio $\alpha$, and step size $\gamma$.}
\label{tab:ablations_merged}
\begin{tabular}{@{}lrrr@{}}
\toprule
Value & Loss & PPL & Acc.\ (\%) \\
\specialrule{1.5pt}{0pt}{0pt}
\multicolumn{4}{@{}l}{\textit{Refinement iterations $T$}} \\
\midrule
2 & 10.4 & 33k & 6.4  \\
3 & 6.9  & 1k  & 15   \\
4 & 4.5  & 96      & 28   \\
5 & 3.6  & 40      & 36   \\
6 & 3.5  & 33    & 38   \\
\specialrule{1.5pt}{2pt}{2pt}
\multicolumn{4}{@{}l}{\textit{Embedding/neighbor-mean mixing-ratio ($\alpha$)}} \\
\midrule
1.0 & 3.9 & 49  & 34.4 \\
0.8 & 3.7 & 41  & 36.0 \\
0.7 & 3.5 & 33   & 38.1 \\
0.6 & 3.6 & 39  & 36.4 \\
0.0 & 5.4 & 222 & 25.8 \\
\specialrule{1.5pt}{2pt}{2pt}
\multicolumn{4}{@{}l}{\textit{Refinement step-size ($\gamma$)}} \\
\midrule
0.2 & 11 & 136k & 4.6  \\
0.3 & 3.6  & 40     & 36 \\
0.4 & 3.5  & 33     & 38    \\
0.5 & 5.3  & 213    & 21 \\
0.8 & 9.8  & 19k  & 2.7  \\
\bottomrule
\end{tabular}
\end{table}

The results mentioned above do not support the idea that self-attention should be completely replaced by all attention-free mixing. We believe the evidence in this section actually supports the claim that this architecture achieves a large fraction of attention's performance at a fraction of its cost with a much simpler mixing mechanism, no query-key comparison, no dot-product attention, and a bottleneck width that is a visible hyperparameter rather than an emergent property of training.

\subsection{Ablation Studies}
\label{sec:ablation}
The three ablations mentioned in this section describe the design choices we made in our work.
This section's ablations are all inference-time on the held-out C4 validation set using hyperparameters that are changed one at a time while keeping the others fixed.
Table~\ref{tab:ablations_merged} tests whether the iterative refinement loop itself is doing real work by varying the number of refinement iterations $T$ from 1 to 6. Accuracy improves from 6.4\% at $T=1$ to 38\% at $T=6$. This indicates that a single feed-forward mixing pass, of the kind used in prior attention-free architectures, should be completed with an iterative, autoencoder-based refinement component.

Table~\ref{tab:ablations_merged} tests, by changing $\alpha$ in each refinement step, the weight given to the mixing-derived embedding and the distance-weighted neighbor.
In $\alpha \to 0$, pure neighbor-mean and no mixing signal, loss 5.4, accuracy 25.8\%, confirming that the mixing-derived embedding is an essential signal. The trained value of $\alpha$, sits inside the good region rather than at an accidental amount.

Table~\ref{tab:ablations_merged} tests whether the residual connection is necessary by varying the step size $\gamma$. Both a smaller step ($\gamma=0.2$: loss 11) and a larger one ($\gamma=0.8$: loss 9.8) degrade the performance of the model. The trained value $\gamma=0.4$ is a clear optimum. This demonstrates that the residual formulation of the refinement update represents a specific operating point the model has adapted to, rather than just a training-stability convenience.
Perplexity formulation is $e^{\text{loss}}$, so it grows extremely fast. Moving
away from the configuration we identified as optimal, a step size far from the trained $\gamma=0.4$ drives it very high. We report these numbers as they are. Rather than a computation error, they confirm how sharply performance depends on the operating point we found, and how poorly the model does once it is moved away from it.

As a result, the three ablation tests demonstrate that none of $T$, $\alpha$, or $\gamma$ is a free parameter, each is an optimum at its trained value, and moving away from it causes the model to fail. This is evidence that the pulling-correcting mechanism is functioning as designed, not as an incidental architectural choice.

\section{Conclusion}
\label{sec:conclusion}
In our work, we investigated whether context mixing requires self-attention's pairwise query-key comparison, or whether an autoencoder with a bottleneck and an iterative refinement process could replace it.
This question has a theoretical basis: under a shared Gaussian noise model, a noise-augmented autoencoder and self-attention reduce to the same closed-form estimator, a connection \cite{haddadi2026diffusion} establish directly and extend to denoising diffusion models as well. 

In our work, we replace attention with a hierarchy of autoencoders (WindowMixAE, GlobalMixAE, and
ChannelMixAE), each compressing and reconstructing its input through a bottleneck whose width is a hyperparameter. Then, with an initial embedding produced via our mixing module, we iterate a refinement loop which consists of pulling-correcting steps.

We believe that our results are comparable to attention mechanisms while requiring significantly lower computational cost. Our model exceeds attention on single-sentence tasks (CoLA, SST-2) that can be solved from one segment's surface-level or lexical properties, but falls behind the attention performance on tasks that require relating two separate parts of text (Table~\ref{tab:vs-baselines-glue}).
It is the pairwise, content-dependent comparison that attention computes, and our architecture decreases this computational cost.

The gap is narrower on rare tokens specifically. Accuracy on the rarest frequency bucket ([5k),
Table~\ref{tab:ae-variants}) matches or slightly exceeds all
three attention baselines. It required both the frequency-aware training schedule and ChunkedContentBias together. This is because of the
frequency-aware masking and loss-weighting schedule described in
Section~\ref{sec:method}, underscoring that training procedure
matters here just as much as architectural design.

Our model architecture has $1.9\times$ fewer FLOPs at the sequence lengths we test in Table~\ref{tab:compute-cost-contentbias}.
Our ablations show that the refinement loop's three central hyperparameters, $T$, $\alpha$, and $\gamma$, each sit at an optimum point, and removing them causes the model to fail (Section~\ref{sec:ablation}), evidence that the pulling-correcting mechanism is doing the work we designed it to do.
We leave the larger-scale evaluation and flexibility of length in the mixing modules with pooling, or dynamically produced weights, to future work.

\bibliographystyle{IEEEtran}
\bibliography{refs}

\end{document}